\documentclass[letterpaper, 10 pt, conference]{ieeeconf}  

\IEEEoverridecommandlockouts                              

\usepackage[T1]{fontenc}

\usepackage{amsfonts}	
\usepackage{amsmath}	
\usepackage{amssymb}    
\usepackage{siunitx}
\usepackage{xfrac}    
\usepackage{pifont}   

\usepackage{booktabs}
\usepackage{makecell}  
\usepackage[flushleft]{threeparttable}  
\usepackage{multirow}
\usepackage{array}

\usepackage{xspace}    
\usepackage[dvipsnames]{xcolor}    
\usepackage{colortbl}
\let\labelindent\relax
\usepackage[inline]{enumitem} 
\usepackage{graphicx} 
\usepackage{microtype}
\usepackage{cite}
\usepackage{flushend}
\usepackage{algorithm2e}

\makeatletter
\let\NAT@parse\undefined
\makeatother

\usepackage{url}

\usepackage[pdfencoding=auto, colorlinks=true]{hyperref} 
\usepackage[hang,flushmargin]{footmisc}

\usepackage[capitalize]{cleveref}
\crefname{section}{Sec.}{Secs.}
\Crefname{section}{Section}{Sections}
\Crefname{table}{Table}{Tables}
\crefname{table}{Tab.}{Tabs.}

\definecolor{graph_teal}{HTML}{006162}
\definecolor{graph_yellow_spheres}{HTML}{E5B931}
\definecolor{graph_yellow_boxes}{HTML}{D5D200}
\definecolor{graph_red}{HTML}{FF3936}
\definecolor{graph_blue}{HTML}{1E22DB}
\definecolor{graph_green}{HTML}{52BA00}

\newcommand{\cmark}{\ding{51}}%
\newcommand{\xmark}{\ding{55}}%

\newcommand{\grayrule}{\arrayrulecolor{black!30}\midrule\arrayrulecolor{black}}

\newcommand{\etal}{\textit{et al.}}

\newcolumntype{x}[1]{>{\centering\arraybackslash\hspace{0pt}}p{#1}}

\newcommand{\method}{\mbox{CURB-OSG}\xspace}

\begin{document}

\title{\LARGE \bf
Collaborative Dynamic 3D Scene Graphs \\ for Open-Vocabulary Urban Scene Understanding
}

\author{
Tim Steinke$^{1*}$, 
Martin Büchner$^{1*}$, 
Niclas Vödisch$^{1*}$, 
and Abhinav Valada$^{1}$
\thanks{$^{*}$ Equal contribution.}%
\thanks{$^{1}$ Department of Computer Science, University of Freiburg, Germany.}%
\thanks{This work was funded by the German Research Foundation (DFG) Emmy Noether Program grant number 468878300.}%
}

\maketitle
\thispagestyle{empty}
\pagestyle{empty}


\begin{abstract}
    Mapping and scene representation are fundamental to reliable planning and navigation in mobile robots. While purely geometric maps using voxel grids allow for general navigation, obtaining up-to-date spatial and semantically rich representations that scale to dynamic large-scale environments remains challenging. In this work, we present CURB-OSG, an open-vocabulary dynamic 3D scene graph engine that generates hierarchical decompositions of urban driving scenes via multi-agent collaboration. By fusing the camera and LiDAR observations from multiple perceiving agents with unknown initial poses, our approach generates more accurate maps compared to a single agent while constructing a unified open-vocabulary semantic hierarchy of the scene. Unlike previous methods that rely on ground truth agent poses or are evaluated purely in simulation, CURB-OSG alleviates these constraints. We evaluate the capabilities of CURB-OSG on real-world multi-agent sensor data obtained from multiple sessions of the Oxford Radar RobotCar dataset. We demonstrate improved mapping and object prediction accuracy through multi-agent collaboration as well as evaluate the environment partitioning capabilities of the proposed approach. To foster further research, we release our code and supplementary material at \mbox{\url{https://ov-curb.cs.uni-freiburg.de}}.

\end{abstract}


\section{Introduction}

Scene graphs have emerged as a prominent representation for modeling complex scenes in robotics. In particular, their integration with semantic 3D maps offers several notable advantages over traditional mapping techniques, such as voxel grids. Most importantly, scene graphs encapsulate high-level topological information, enabling efficient information retrieval for a range of downstream tasks, including navigation~\cite{greve2024curbsg,hovsg, buchner2023learning} and manipulation~\cite{honerkamp2024language}. Prior studies have highlighted these benefits in both indoor~\cite{hovsg} and outdoor~\cite{greve2024curbsg, deng_opengraph_2024} environments, emphasizing the broad applicability of this approach.
\looseness=-1

A key aspect in the construction of scene graphs is the detection of objects within the environment whose categories can be drawn from a closed vocabulary~\cite{hughes2022hydra} or using open-vocabulary perception~\cite{hovsg, gu_conceptgraphs_2024}, paving the way toward open-world deployment. To improve mapping accuracy and topicality, recent scene graph construction methods~\cite{greve2024curbsg, chang_hydra-multi_2023} incorporate inputs from multiple agents. These collaborative approaches not only facilitate information sharing among multiple agents but also support frequent updates to the scene graph. However, merging objects detected by different agents into a consistent representation remains a highly challenging task, particularly in the absence of accurate global agent poses.
\looseness=-1

In previous work, we proposed constructing multi-agent dynamic 3D scene graphs in large-scale outdoor driving scenes~\cite{greve2024curbsg}. Similar to other methods~\cite{chang_hydra-multi_2023}, the approach is constrained by closed-set semantics and was evaluated with a photorealistic simulator~\cite{DosovitskiyCarla17}. In addition, its collaborative SLAM backend relies on global agent pose initialization, which eases the fusion of object proposals under mapping ambiguity.
\looseness=-1


\begin{figure}[t]
    \centering
    \includegraphics[width=\linewidth]{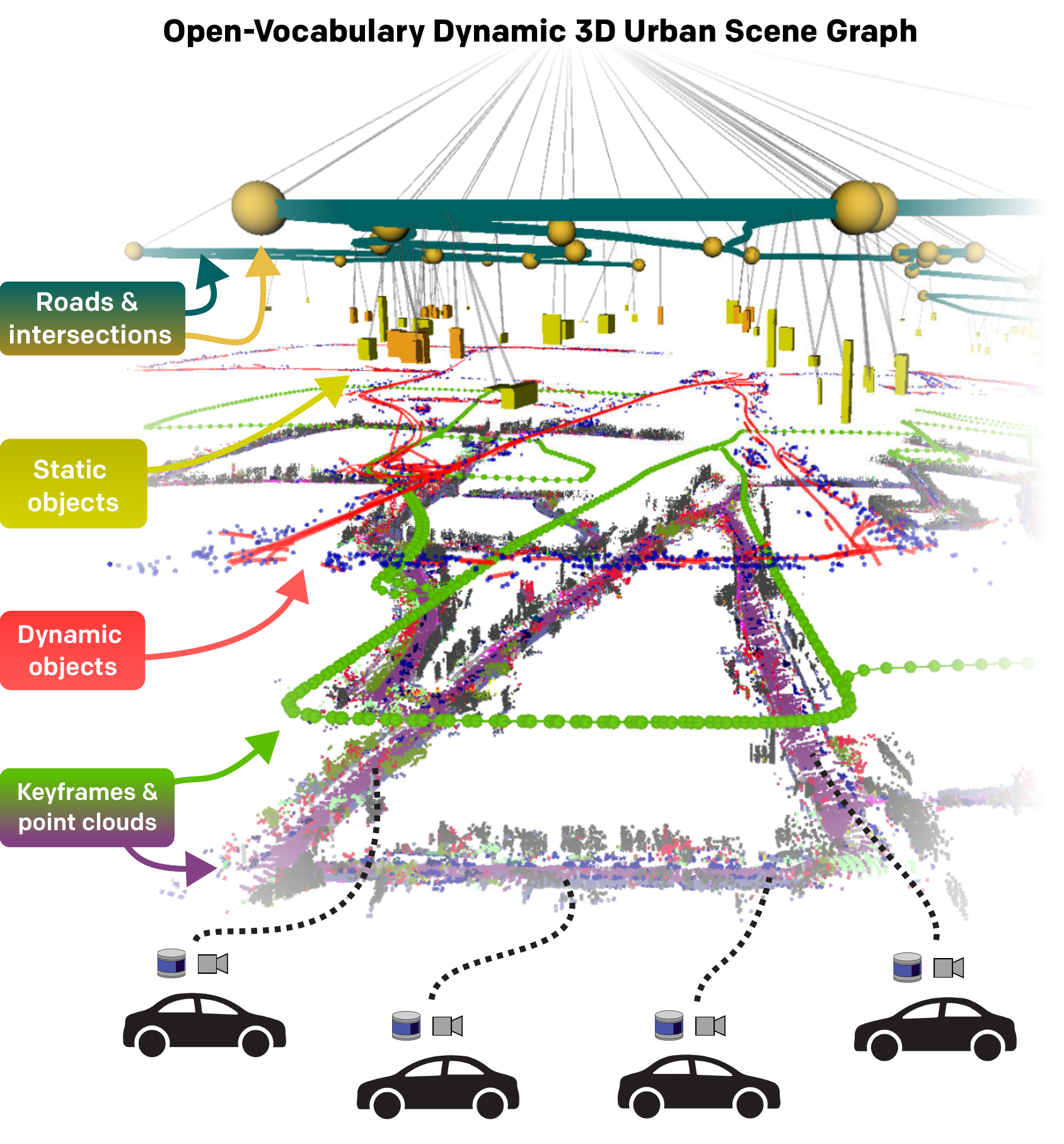}
    \vspace*{-.8cm}
    \caption{We present \method for generating open-vocabulary collaborative dynamic 3D scene graphs to model urban driving scenes. We fuse the observations of multiple agents by performing global inter-agent graph optimization using a centralized mapping instance without assuming initial agent poses. A tight coupling with a 3D scene graph engine allows for merging object proposals using open-vocabulary semantics even under mapping ambiguities.
    \looseness=-1
    }
    \label{fig:teaser}
    \vspace*{-.5cm}
\end{figure}


In this work, we address these limitations by extending our prior research to \textit{Collaborative URBan Open-vocabulary Scene Graphs} (\method) and investigating its application in real-world urban driving scenes involving multiple agents.
Our \method approach integrates a collaborative graph SLAM backend with global inter-agent loop closure detection, eliminating the need for global pose initialization and enabling centralized construction of a consistent 3D map. Furthermore, \method incorporates open-vocabulary perception, running on each agent and allowing for querying scene-specific objects. Finally, we construct a large-scale scene graph, as shown in \cref{fig:teaser}, that covers a combined travel distance of \qty{28}{\km} and automatically extracts road intersections, static objects such as traffic lights, and other dynamic road users.

To summarize, our main contributions are as follows:
\begin{enumerate}[topsep=0pt]
    \item We introduce \method, a scene graph engine that features a collaborative SLAM backend, which integrates inter-agent loop closures and operates without initial agent pose estimates.
    \item We extend our previous scene graph framework to open-vocabulary semantics, allowing operation in real-world urban driving scenes.
    \item We develop a novel test bed for multi-agent urban mapping based on the Oxford Radar RobotCar Dataset~\cite{barnes2020robotcarradar}.
    \item We conduct comprehensive evaluations of \method to analyze its spatial and semantic capabilities in the face of multi-agent ambiguities.
    \item We make our code and sample data publicly available at \mbox{\url{https://ov-curb.cs.uni-freiburg.de}}.
\end{enumerate}

\section{Related Work}
\label{sec:related-work}

We provide an overview of LiDAR SLAM, multi-agent variants, and recent approaches for generating scene graphs.


{\parskip=2pt
\noindent\textit{LiDAR SLAM:}
LiDAR-based odometry~\cite{zhang2014loam,vizzo2023kissicp} and mapping~\cite{koide2019hdlgraphslam} methods are widely utilized in robotics due to their precise pose estimates and detailed 3D models.
In simultaneous localization and mapping (SLAM), loop closure detection is a key aspect that can be performed at either a local or global scale.
When odometry drift is minimal, loop closure is typically performed by registering point clouds with potential candidates and accepting the result if a "fitness score" exceeds a threshold. However, this approach becomes impractical for large-scale comparisons. To address this problem, global loop closure detection prefilters candidates using scan descriptors.
One commonly used descriptor is Scan Context~\cite{kim2018scancontext}, which encodes low-level features such as point coordinates and intensities. At test time, a search algorithm identifies locations with similar descriptors, substantially reducing the number of loop closure candidates.
Other descriptors incorporate more abstract features, e.g., derived from semantic segmentation~\cite{li2021semanticscancontext} or visual foundation models~\cite{voedisch2025vfmreg}.
Alternatively, some approaches address global loop closure detection in an end-to-end manner~\cite{arce2023padloc, cattaneo2022lcdnet}, bypassing the two-step process of scan encoding and candidate search.
However, a key limitation of these methods is their dependence on domain-specific training.
In this work, we build upon our previous adaptations~\cite{greve2024curbsg} to HDL Graph SLAM~\cite{koide2019hdlgraphslam}. Specifically, we employ Scan Context~\cite{kim2018scancontext} to enable loop closure registration between different agents without global pose initialization.
}


{\parskip=2pt
\noindent\textit{Collaborative SLAM:}
To enhance scalability and mapping efficiency, various studies have integrated multi-agent cooperation into the SLAM problem~\cite{zou2019collaborative}.
Centralized approaches aggregate data from all agents on a central server to construct a consistent map. In contrast, distributed approaches employ decentralized algorithms and peer-to-peer communication, enabling agents to share computational load.
While early research focused on centralized visual-inertial SLAM~\cite{riazuelo2014c2tam, Karrer2018cvislam}, subsequent works have extended this concept to LiDAR-based methods. For instance, our previous work CURB-SG~\cite{greve2024curbsg} employs a centralized framework for semantic mapping, while LAMP~2.0\cite{chang2022lamp2} enables large-scale multi-agent collaboration in underground environments. For a similar use case, Swarm-SLAM~\cite{lajoie2024swarmslam} adopts a distributed approach integrating multiple sensor modalities.
A key challenge in multi-agent SLAM is inter-agent loop closure detection~\cite{huang2022discoslam}, especially without additional information such as GNSS. In this work, we specifically address this limitation of our previous approach~\cite{greve2024curbsg} via a global loop closure search between the agents.\looseness=-1
}


{\parskip=2pt
\noindent\textit{3D Scene Graphs:}
3D scene graphs constitute structured, metric representations of 3D environments, encoding the spatial and semantic relationships of scene entities such as objects or regions~\cite{hughes2022hydra}. Their inherent sparsity enables efficient space partitioning, facilitating faster inference compared to dense representations. As a result, they serve as an effective interface between robotic maps generated from LiDAR or camera data and downstream tasks such as navigation~\cite{greve2024curbsg, hovsg} and manipulation~\cite{honerkamp2024language}.
Various works have applied this principle to indoor mapping~\cite{armeni20193d, hughes2022hydra, sgraphs_2022}, leveraging closed-set semantic perception pipelines. Only a few approaches achieve real-time capability~\cite{wu_scenegraphfusion, sgraphs_2022} or can handle dynamic scenes~\cite{hughes2022hydra}.
A newer branch of work addresses the limitation of closed-vocabulary semantics by incorporating open-vocabulary perception modules to represent indoor scenes using flexible category definitions~\cite{hovsg, gu_conceptgraphs_2024}. Traditionally, the task of inferring object instances based on image data has been approached by training models on labeled datasets with fixed object categories~\cite{voedisch2025pastel, redmonYouOnlyLook2016} providing high accuracy while being limited to categories seen during training time. More recently, advances in vision-language pretraining~\cite{liGroundedLanguageImagePretraining2022,liBLIPBootstrappingLanguageImage2022a} have enabled open-vocabulary semantic object understanding. Coupled with large-scale segmentation models~\cite{kirillovSegmentAnything2023}, several approaches are capable of open-vocabulary instance and semantic understanding~\cite{liuGroundingDINOMarrying2024, renGroundedSAMAssembling2024, zhangRecognizeAnythingStrong2024}. In this work, we incorporate open-vocabulary vision-language models into our scene graph mapping engine.}

Only a small share of works have investigated the applicability to outdoor scenes. Strader~\etal~\cite{strader_indoor_2024} build scene graphs of arbitrary environments using large language models and identify plausible concept relations using logic tensor networks. A limitation of the approach is its reliance on preconstructed maps. 
Our previous CURB-SG~\cite{greve2024curbsg} introduces a dynamic 3D scene graph construction engine targeting urban driving scenes. It decomposes urban environments into roads and intersections while including static objects such as traffic lights and dynamic entities such as cars. The navigational topology is represented by a lane graph layer. Importantly, CURB-SG aggregates observations from multiple agents.
Orthogonally, OpenGraph~\cite{deng_opengraph_2024} proposes an offline approach for constructing hierarchical 3D scene graphs of driving scenes using open-vocabulary vision-language models. The proposed representation contains a road graph layer and a segments layer, where segments are contiguous clusters of visited places. Nonetheless, OpenGraph relies on given poses and is, as such, not tightly coupled with a mapping approach.
In this work, we construct dynamic 3D scene graphs of outdoor driving scenes through multi-agent cooperation and leverage open-vocabulary vision-foundation models.
\begin{figure*}[t]
    \centering
    \includegraphics[width=1\textwidth]{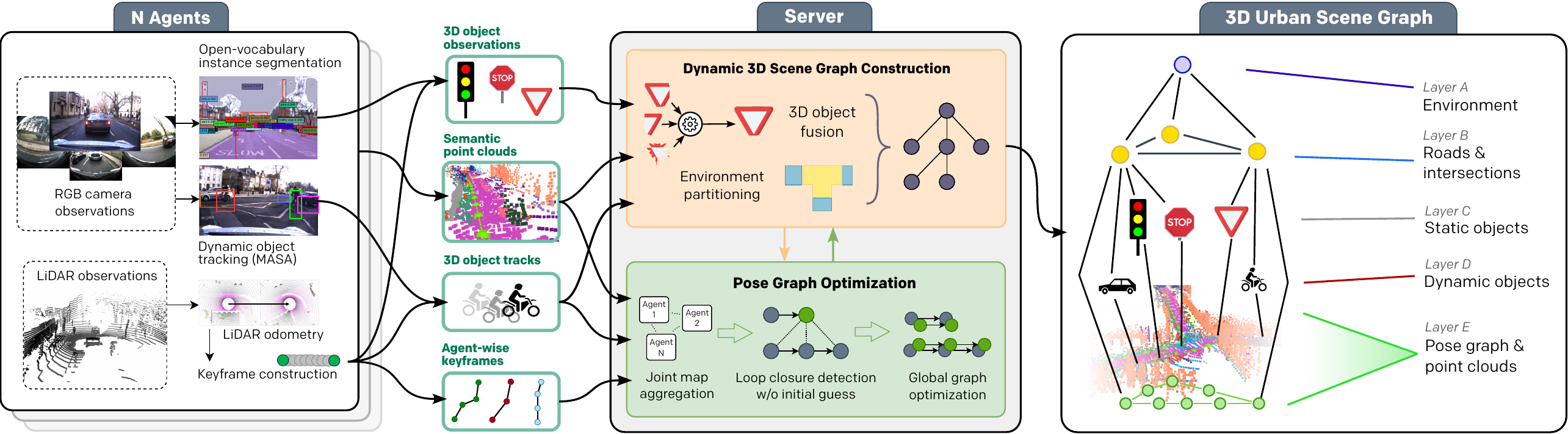}
    \vspace*{-.4cm}
    \caption{An overview of our \method approach operating on LiDAR and camera data from multiple agents. On each agent, we perform open-vocabulary perception that processes monocular image data from surround-view cameras to extract 2D object detections using Grounding DINO~\cite{liuGroundingDINOMarrying2024} and dynamic object tracks through MASA~\cite{liMatchingAnythingSegmenting2024}. Simultaneously, we estimate each agent's LiDAR odometry via scan matching and construct keyframes that are sent to a central server. All object observations, both static and dynamic, are projected onto the filtered LiDAR point clouds, extracted to obtain 3D object observations relative to the keyframe poses. The central server receives the keyframes and runs graph-based SLAM~\cite{koide2019hdlgraphslam} coupled with LiDAR-based loop closure detection~\cite{kim2018scancontext} to estimate a joint pose graph holding the historic poses of all agents. Finally, all object observations and the semantic point clouds are processed in our 3D scene graph construction module to obtain a unified, hierarchical representation of the environment.}
    \label{fig:overview}
    \vspace*{-.2cm}
\end{figure*}


\section{Technical Approach}

We illustrate the modular architecture of our proposed \method approach in \cref{fig:overview}. \method follows a centralized collaboration model, where data from several agents is aggregated in a central server. In \cref{sec:approach-cslam}, we describe our approach for collaborative SLAM. The agents transmit their LiDAR data and odometry estimates to the server, which performs intra- and inter-agent loop closure detection and pose graph optimization to build a unified map of the environment. In \cref{sec:approach-perception}, we present our perception approach that runs on the agents and outputs 3D observations that are used to construct the scene graph. We leverage open-vocabulary vision-language models to segment static and dynamic objects in monocular images and project them onto the LiDAR point clouds. In \cref{sec:approach-scene-graph}, we outline our scene graph module, which efficiently processes the optimized pose graph and the agent observations into a globally consistent hierarchical 3D scene graph. The module dynamically fuses 3D observations into nodes, forming layers for static and dynamic objects. It also partitions the optimized pose graph into roads and intersections to construct an intermediate road graph layer.
\looseness=-1


\subsection{Collaborative SLAM}
\label{sec:approach-cslam}

We use collaborative LiDAR SLAM to construct a pose graph as the geometrical foundation of our scene graph. We build on top of HDL Graph SLAM~\cite{koide2019hdlgraphslam}, extending it to handle multiple agents with unknown initial alignment.


{\parskip=2pt
\noindent\textit{Agents:}
The agents are equipped with 3D LiDAR sensors to capture point clouds of the environment. Each agent initially filters the point clouds to remove outliers, followed by a downsampling step. Next, it applies pairwise scan matching to subsequent point clouds with Fast~GICP~\cite{koideVoxelizedGICPFast2021} to obtain LiDAR odometry estimates. Using the odometry, the agent generates keyframes at regularly spaced intervals to reduce the amount of data transmitted. Finally, the keyframes are sent to the central server, including pose estimates, agent IDs, and the downsampled LiDAR scans.


\noindent\textit{Server:}
The server aggregates keyframes from all agents into a single pose graph. Upon receiving a new keyframe, HDL Graph SLAM~\cite{koide2019hdlgraphslam} adds it to the pose graph and searches for loop closure candidates within a pre-defined radius of the estimated pose. In contrast to previous work~\cite{greve2024curbsg}, the initial global alignment between the agents is unknown, so we cannot rely on odometry estimates to limit the search for inter-agent loop closures. Instead, we leverage ScanContext~\cite{kim2018scancontext} point cloud descriptors to efficiently search for global loop closure candidates. The server validates loop closure candidates by scan matching and adds them to the pose graph if the residual error is below a predefined threshold. Finally, the server performs pose graph optimization~\cite{kuemmerle2011g2o} to create a consistent 3D map. To address mapping problems caused by dynamic objects in the environment, we employ a vision-language model as described in \cref{sec:approach-perception} and remove dynamic objects from the point clouds before searching for loop closures.
}


\subsection{Open-Vocabulary Perception}
\label{sec:approach-perception}
In this section, we describe our object perception approach, which uses open-vocabulary vision-language models to perceive static and dynamic objects. Our proposed modules process monocular images and project their results onto the keyframe point clouds to generate 3D observations, further compressing the sensor information to save bandwidth. We first describe our perception method for static objects, followed by dynamic object tracking.

\begin{figure}[t]
    \centering
    \includegraphics[width=\linewidth]{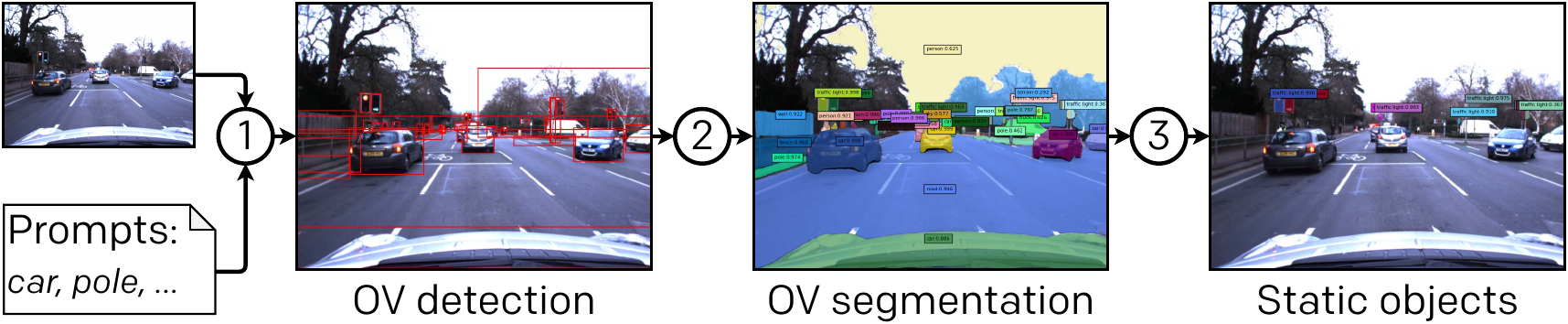}
    \caption{In our open-vocabulary (OV) perception module, (1)~we employ Grounding DINO\cite{liuGroundingDINOMarrying2024} to detect relevant semantic categories, (2)~utilize these detections as prompts for TAP\cite{pan2024tokenizeAnything} to generate semantic masks, and (3)~filter the output to include only objects represented in the scene graph.}
    \label{fig:opengraph-our-pipeline}
\end{figure}

{\parskip=2pt
\noindent\textit{Static Objects and Semantic Point Clouds:}
For our static object perception approach, we advance the open-vocabulary ensemble method of OpenGraph~\cite{deng_opengraph_2024} by reducing its size and capabilities to allow for real-time inference. As illustrated in \cref{fig:opengraph-our-pipeline}, we first prompt the Grounding DINO~\cite{liuGroundingDINOMarrying2024} detector with a set of classes relevant to urban driving scenes~\cite{cordts2016cityscapes}. Since this set can be selected according to the specific use case, this procedure does not limit our open-vocabulary approach. Next, we extend these bounding box-based detections to segmentation masks using TAP~\cite{pan2024tokenizeAnything} and filter the output to include only objects captured in the scene graph.
\looseness=-1
}

We then reproject the object masks onto the point clouds for two purposes: First, we construct a semantic point cloud layer with point-wise class annotations and, second, we extract the points of relevant static objects such as traffic lights and signs. Since we observed that these static object points contain a considerable number of outliers due to reprojection errors and mask inaccuracies, we employ a density and range-based outlier rejection scheme. This filter excludes points too far in front or behind the median of the object points. The resulting 3D objects are transmitted to the server, where they are fused to build the static objects layer of our scene graph. We describe this step in \cref{sec:approach-scene-graph}.


\begin{figure}[t]
    \centering
    \includegraphics[width=\linewidth]{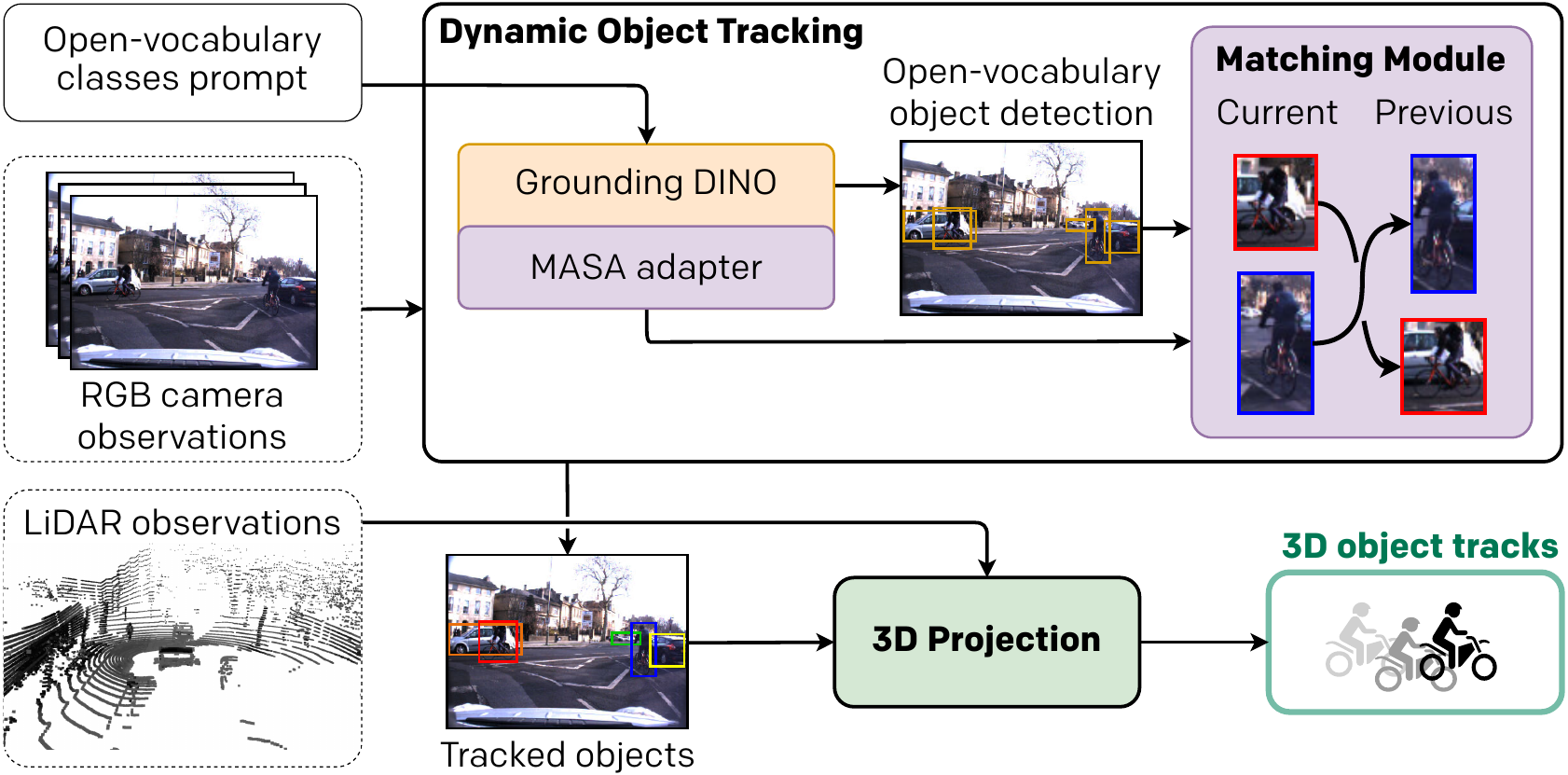}
    \caption{Our dynamic object perception module uses MASA~\cite{liMatchingAnythingSegmenting2024} with the Grounding DINO~\cite{liuGroundingDINOMarrying2024} detector to track objects across sequential images. We project the detections onto the point cloud to obtain dynamic 3D object observations. These are then transmitted to the server, contributing to the dynamic objects layer.}
    \label{fig:masa-sam-pipeline}
\end{figure}

{\parskip=2pt
\noindent\textit{Dynamic Object Tracking:}
We show our dynamic object tracking module in \cref{fig:masa-sam-pipeline}. Specifically, we use Grounding DINO~\cite{liuGroundingDINOMarrying2024} for open-vocabulary object detection and rely on MASA~\cite{liMatchingAnythingSegmenting2024} for 2D object tracking in sequential camera images. The so-called MASA adapter extracts visual features that aid the matching module in finding valid associations with previously seen objects to output temporally consistent instance IDs. Each agent projects the 2D detections onto its corresponding point cloud, yielding the temporal 3D observations that are transmitted to the server. As detailed in \cref{sec:approach-scene-graph}, the server uses the instance IDs to generate tracks for the dynamic objects layer of our scene graph.
}


\subsection{3D Scene Graph Generation}
\label{sec:approach-scene-graph}

The scene graph module is central to our approach, integrating the observations with the optimized pose graph to generate a dynamic, globally consistent 3D scene graph. Our scene graph, which is depicted in \cref{fig:overview}, comprises multiple layers:
\begin{enumerate*}[label={(\Alph*)}]
    \item The root layer references the entire environment.
    \item This layer comprises roads and intersections of the scene.
    \item The static objects layer contains the fused results of the agents' static object observations.
    \item The dynamic objects layer is built from the agents' object tracking observations.
    \item The final layer is the optimized 3D map, including the pose graph from collaborative SLAM and the annotated point clouds from the agents.
\end{enumerate*}

In this section, we first describe the construction of the road and intersection nodes, then explain our method for fusing static object observations, and finally, we outline the generation of dynamic object tracks.


{\parskip=2pt
\noindent\textit{Roads and Intersections:}
For our road graph layer, we adopt a heuristic approach. In contrast to previous work~\cite{deng_opengraph_2024}, we consider multi-agent input. Specifically, our approach uses a measure of \textit{path disfluency} to identify sharp turns in an agent's trajectory and clusters these points with DBSCAN~\cite{ester1996dbscan} to obtain intersection nodes. The edges are inferred from the agent trajectories connecting these intersections and subsequently de-duplicated to estimate the road topology. While this method is fast and robust, a limitation is that it cannot discern between intersections and sharp turns in the road and only detects intersections where an agent has changed direction.
}


{\parskip=2pt
\noindent\textit{Static Object Fusion:}
Our scene graph module aggregates and fuses static object observations. First, upon receiving a new observation from an agent, the module uses the optimized pose graph to transform the observation into a global frame. Next, it attempts to associate the observation with an existing object node by considering its semantic class and the bounding box overlap. We manually tune a threshold overlap ratio to reduce false associations and duplicated objects.
Besides integrating new observations, the scene graph module must also dynamically react to pose graph updates. We achieve this by re-transforming the observations whenever a large change is detected, after which we check the consistency of the affected objects. We do this by re-computing the bounding box overlap to detect newly valid and invalid associations, followed by merging and splitting the object nodes as needed.
While these operations are computationally expensive for large environments, they are required to maintain accuracy and consistency with the rest of the scene graph, and our efficient implementation ensures reasonable performance. 
}


{\parskip=2pt
\noindent\textit{Dynamic Object Tracks:}
We combine the 3D observations of the agents into global dynamic object tracks. Since each observation is relative to an agent's local frame, we leverage the optimized pose graph to estimate the global position at the time of observation. We then combine multiple observations of the same object associated with MASA~\cite{liMatchingAnythingSegmenting2024} to obtain a global movement track. To mitigate false associations, we split the track at points where two subsequent observations are spaced more than \qty{30}{\m} apart. In the next step, we distinguish between still-standing dynamic objects, e.g., parked cars, from objects in motion by applying a threshold of \qty{8}{m} traveled from the initially observed position. Finally, we add both dynamic object types to the scene graph and link them to the road or intersection where they were observed.
}

\section{Experimental Evaluation}

In this section, we introduce our real-world multi-agent experimental setting~(\cref{sec:multi-robotcar}), followed by evaluating the main components of \method, namely multi-agent mapping~(\cref{sec:exp-mapping}), open-vocabulary object fusion~(\cref{sec:exp-object-fusion}), and the predicted high-level topology~(\cref{sec:exp-topology}).


\subsection{Multi-Agent Dataset}
\label{sec:multi-robotcar}

The Oxford Radar Robotcar Dataset~\cite{barnes2020robotcarradar} is a large-scale outdoor dataset consisting of sensor data from an extensive sensor suite built atop a car platform. The dataset contains 32 sequences recorded on seven different days. It features radar, 2D and 3D LiDAR, stereo and monocular cameras, and combined GNSS/INS data. Throughout the evaluation, we make use of the provided $SE(2)$ pose ground truth that is generated through multi-sensor pose graph optimization across the sequences and is thus less noisy compared to the 3D INS ground truth. We only draw data from the RGB cameras and the 3D LiDARs as input to our approach.

As part of this work, we contribute a ROS-based Radar RobotCar Dataset player\footnote{Open-sourced multi-agent Oxford Radar RobotCar player: \\ \url{https://github.com/TimSteinke/multi_robotcar}} that can play multiple different sequences in parallel, thus simulating a multi-agent exploration setup. Furthermore, the tool allows for delayed playback of single sequences and clipping in order to \textit{launch} an agent, e.g., in the middle of its corresponding sequence. Based on this, we can replicate a collaborative mapping scenario using real-world data as the sequences predominantly cover identical routes. 
For the experiments, we randomly choose 3 out of the 32 provided sequences, namely \texttt{2019-01-11 13:24:51}, \texttt{2019-01-14 14:15:12}, and \texttt{2019-01-15 13:06:37}.

Irrespective of the number of agents considered per experiment, we average the mapping results across all possible combinations of agents to rule out deviations stemming from the non-deterministic behavior of the loop closure detection module. In multi-agent scenarios, we begin the runs of agents 2 and 3 at a later timestamp within the respective sequence to simulate different initial positions.


\subsection{Localization and Mapping}
\label{sec:exp-mapping}

In the initial experiment, we evaluate the collaborative SLAM approach of our proposed \method with respect to the ground truth poses provided by the Radar RobotCar Dataset~\cite{barnes2020robotcarradar}. Note that the ground truth data was generated using 2D visual-radar odometry, followed by joint pose graph optimization across all sequences. Consequently, we conduct our evaluation in the 2D domain as well. To ensure alignment with the ground truth coordinate frame, we initialize the first pose of agent 1 with the corresponding GNSS/INS pose.
We report the absolute trajectory error~(ATE)~\cite{filipenko2018comparison} as a measure of overall mapping performance via aligning position estimates based on timestamps and comparing against ground truth positions.
In addition, we report the relative translational~(E\textsubscript{trans}) and rotational odometry errors~(E\textsubscript{rot})~\cite{geiger2013vision} that quantify local mapping quality while not being affected by global map drift.
\looseness=-1

\begin{table}[t!]
\centering
\caption{Localization and Mapping Performance}
\vspace{-0.2cm}
\label{tab:mapping-eval}
\setlength\tabcolsep{7pt}
\begin{threeparttable}
    \begin{tabular}{l | c | ccc}
        \toprule
        \makecell[c]{\textbf{Agent}} & \textbf{Dyn. object} & \textbf{ATE} & \textbf{E\textsubscript{trans}} & \textbf{E\textsubscript{rot}} \\
        \makecell[c]{\textbf{count}} & \textbf{removal} & [\unit{\m}] & [\unit{\percent}] & [\unit{\deg\per\km}] \\
        \midrule
        \multirow{2}{*}{1 agent}  & \xmark & 17.8$\pm$11.6 & 2.2$\pm$2.8 & 7.7$\pm$13.4 \\
                                  & \cmark & 21.7$\pm$14.9 & 2.3$\pm$2.8 & 7.5$\pm$12.6 \\
        \grayrule
        \multirow{2}{*}{2 agents} & \xmark & \underline{17.0}$\pm$11.4 & \underline{2.0}$\pm$2.6 & \underline{6.8}$\pm$12.1 \\
                                  & \cmark & 17.4$\pm$11.9 & 2.1$\pm$2.6 & \underline{6.8}$\pm$11.9 \\
        \grayrule
        \multirow{2}{*}{3 agents} & \xmark & \textbf{15.7}$\pm$10.2 & \textbf{1.1}$\pm$1.0 & \textbf{3.4}$\pm$4.7\phantom{0} \\
                                  & \cmark & 17.3$\pm$11.3 & 2.2$\pm$2.5 & 7.1$\pm$11.6 \\
        \bottomrule
    \end{tabular}
    Evaluation of the mapping and odometry performance of our collaborative SLAM approach throughout exploration in scenarios with one, two, and three agents. We provide the mean and the standard deviation across the agents and averaged over time, detailed in \cref{fig:mate-plot}. Bold and underlined mean values indicate the best and second-best scores, respectively.
\end{threeparttable}
\end{table}

\begin{figure}[t]
    \centering
    \includegraphics[width=1.0\linewidth]{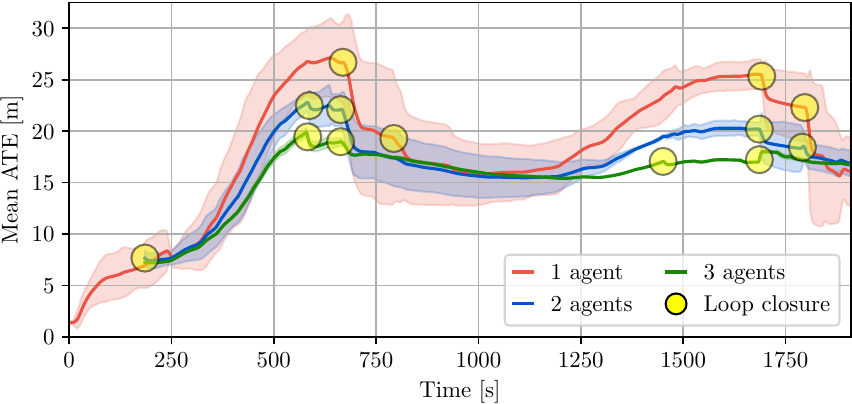}
    \vspace*{-.7cm}
    \caption{Evolution of the mean absolute trajectory error (ATE) over time without dynamic object removal. We provide the mean and the standard deviation across the agents. The temporal average of the ATE is reported in \cref{tab:mapping-eval}. In the case of multi-agent mapping, we plot the ATE as soon as an initial loop closure is incorporated, thus appearing delayed.}
    \label{fig:mate-plot}
    \vspace*{-.4cm}
\end{figure}

We consider six different configurations ranging from one to three agents and evaluate whether dynamic object removal via multi-object tracking (see \cref{sec:approach-perception}) aids the mapping quality. In \cref{tab:mapping-eval}, we report the errors temporally averaged over the map exploration phase. In the multi-agent experiments, we consider the errors starting after the first loop closure to ensure a consistent coordinate frame.
In general, we observe that the overall mapping error is lowest under maximal collaboration (3~agents) as measured using the ATE. Similarly, we also note that the relative odometry errors decrease as more agents are considered. We conclude that the precision of the loop closures incorporated into the multi-agent pose graph is sufficient, enhancing rather than hindering the overall mapping process and effectively leveraging collaboration. Furthermore, we observe that the dynamic object removal does not improve mapping quality, similar to the findings of SUMA++\cite{chen2019iros}. We attribute this to non-mature multi-object tracking using MASA~\cite{liMatchingAnythingSegmenting2024}, which is executed at a lower frequency than the remaining components of our pipeline due to its considerable inference time.
Additionally, we visualize the mapping error throughout the map exploration in \cref{fig:mate-plot}. We observe that the three-agent configuration consistently yields the smallest ATE, improving accuracy compared to the single-agent configuration. The drops, highlighted as yellow circles, are due to the detection of loop closures and the subsequent pose graph optimization.


\subsection{Open-Vocabulary Object Proposals}
\label{sec:exp-object-fusion}

\begin{table}[t!]
\centering
\caption{Static Object Reprojection}
\vspace{-0.2cm}
\label{tab:landmark-reproj-eval}
\setlength\tabcolsep{2.0pt}
\begin{threeparttable}
    \begin{tabular}{l|ccc}
        \toprule
        \textbf{Method} & \textbf{mIoU} [\unit{\percent}] & \textbf{Precision} [\unit{\percent}] & \textbf{Recall} [\unit{\percent}] \\
        \midrule
        Ours w/ MaskCLIP~\cite{zhou2022maskclip} & 27.01 & 68.29 & \phantom{0}8.54 \\
        Ours w/ Grounding DINO~\cite{liuGroundingDINOMarrying2024} & \textbf{31.83} & \textbf{88.22} & \textbf{14.92} \\
        \bottomrule
    \end{tabular}
    Evaluation of static objects through reprojection into image space, utilizing 100 human-annotated images for assessment.
\end{threeparttable}
\end{table}

\begin{table}[t!]
\centering
\caption{Static Object Fusion}
\vspace{-0.2cm}
\label{tab:landmark-fusion}
\begin{threeparttable}
    \begin{tabular}{l| x{.7cm}x{.6cm} |ccc}
        \toprule
        \makecell[c]{\textbf{Agent}} & \multicolumn{2}{c|}{\textbf{Observed by}} & \textbf{mIoU} & \textbf{Precision} & \textbf{Recall} \\
        \makecell[c]{\textbf{count}} & any & all & [\unit{\percent}] & [\unit{\percent}] & [\unit{\percent}] \\
        \midrule
        \multirow{1}{*}{1 agent}  & \cmark & \cmark & 30.68 & 86.23 & 18.27 \\
        \grayrule
        \multirow{2}{*}{2 agents} & \cmark &        & 28.88 & 82.11 & \textbf{25.25} \\
                                  &        & \cmark & 29.31 & \underline{91.67} & 14.75 \\
        \grayrule
        \multirow{2}{*}{3 agents} & \cmark &        & \underline{31.27} & 71.22 & \underline{24.18} \\
                                  &        & \cmark & \textbf{38.15} & \textbf{100.00}\phantom{0} & 10.06 \\
        \bottomrule
    \end{tabular}
    Evaluation of the impact of multi-agent collaboration on the open-vocabulary detection of static objects.    
\end{threeparttable}
\vspace{-0.2cm}
\end{table}

In the following, we analyze the object fusion capabilities of CURB-OSG. Since multiple agents induce a significant amount of ambiguity and noise into the collaborative pose graph, fusing object proposals detected in either one of the agents (see \cref{fig:overview}) with proposals stemming from other agents is a non-trivial task. This complexity arises not only from map offsets and potential missing loop closures but also from the limited resolution of the underlying LiDAR point clouds, ultimately impeding the \textit{anchoring} and re-identification of 2D object proposals within 3D point clouds. In the following, we conduct an ablation study on different perception backbones for static object detection and evaluate to what degree collaboration aids the precision and recall of detected objects. To evaluate the semantic elements of our produced scene graph hierarchy, we have manually attained panoptic labels for all \textit{thing} objects across 100 diverse camera observations following the category convention introduced in the Cityscapes dataset~\cite{cordts2016cityscapes}. Similar to the sequence player, we publicly release these labels as part of our work.

We compare two different perception backbones: In addition to the approach detailed in \cref{sec:approach-perception}, we propose an efficient baseline method utilizing MaskCLIP~\cite{zhou2022maskclip} and FeatUp~\cite{fufeatup} for up-projection to obtain dense semantic image features. In order to obtain pixel-wise class predictions, we encode the set of considered classes in an open-vocabulary manner and score them against the image features using cosine similarities. Next, we group pixels of the same class using connected components labeling~\cite{he2017connectedComponents} to obtain open-vocabulary instance segmentation masks. In \cref{tab:landmark-reproj-eval}, we compare this baseline against our proposed method. We opt for an evaluation in the 2D camera domain due to the lack of 3D ground truth and utilize typical object detection metrics. Under reprojection, we observe vastly increased precision and recall of detected objects using the proposed TAP~\cite{pan2024tokenizeAnything} + Grounding DINO~\cite{liuGroundingDINOMarrying2024} pipeline as well as substantial increases in the mean intersection over union~(mIoU) metric. However, the Grounding DINO-based pipeline is considerably more compute-intensive than the MaskCLIP baseline. Nonetheless, we note that by utilizing a variably defined open-vocabulary set of object categories, we greatly improve inference times of the perception module compared to previous work relying on image tagging methods~\cite{deng_opengraph_2024}. 

In \cref{tab:landmark-fusion}, we additionally evaluate the impact of different multi-agent configurations on open-vocabulary detection. We find that, similar to the mapping results, a higher number of agents increases object detection performance in terms of mIoU as well as precision and recall. For three agents, we find that the proposed pipeline yields a precision of \qty{100}{\percent}, underlining the aspect of certainty in its predictions under collaborative perception. In general, we observe only moderate recalls due to LiDAR-to-camera reprojection errors and far-away objects that are not detected.


\begin{figure*}[t]
    \centering
    \includegraphics[width=0.9\linewidth]{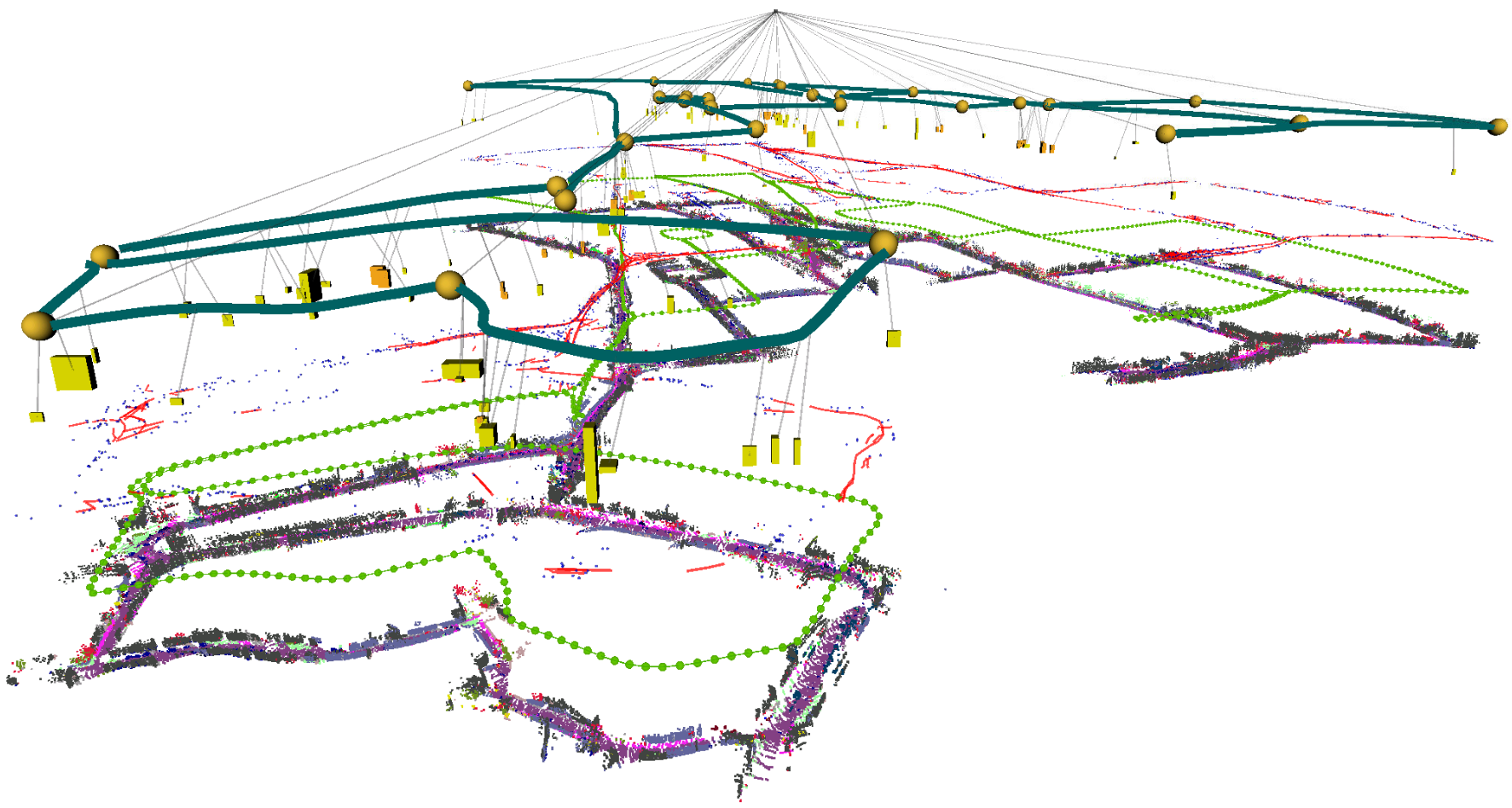}
    \caption{Collaborative open-vocabulary 3D scene graph hierarchies produced by CURB-OSG: The layers from top to bottom represent high-level road graph topology (\textcolor{graph_teal}{teal}) including intersections (\textcolor{graph_yellow_spheres}{yellow spheres}), 3D object proposals anchored to their particular high-level location (\textcolor{graph_yellow_boxes}{yellow boxes}), dynamic object tracks (\textcolor{graph_red}{red}), tracked static objects (\textcolor{graph_blue}{blue}), collaborative pose graph (\textcolor{graph_green}{green}), and the generated semantic 3D map.}
    \label{fig:scene-graph-viz}
\end{figure*}


\subsection{High-Level Topology}
\label{sec:exp-topology}

\begin{figure}
    \centering
    \fbox{\includegraphics[width=.8\linewidth]{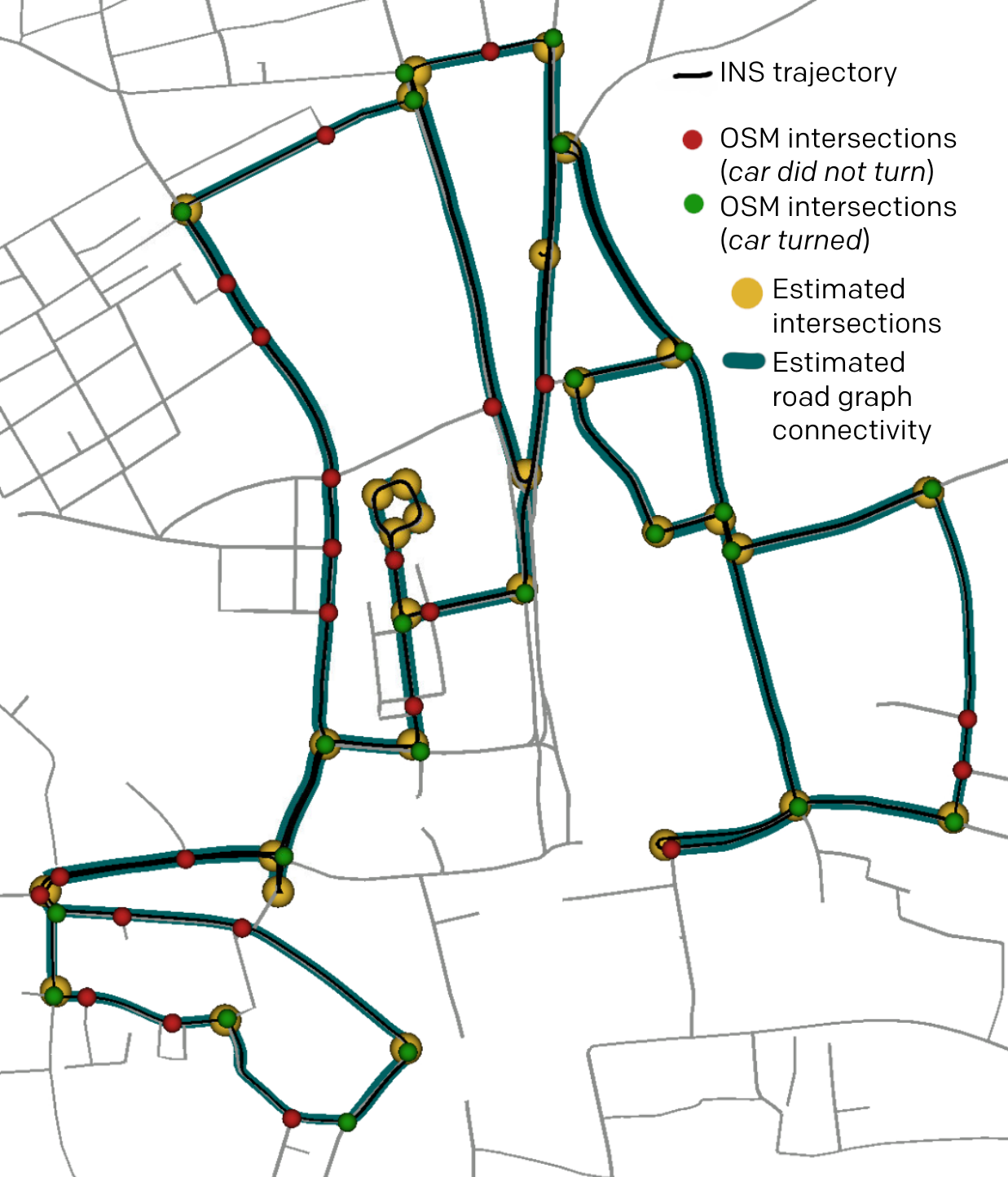}}
    \caption{The road graph from the single-agent evaluation overlayed on the ground truth data from OpenStreetMap.}
    \label{fig:osm-gt}
    \vspace*{-.5cm}
\end{figure}

\begin{table}[t!]
\centering
\caption{Detection of Road Intersections}
\vspace{-0.2cm}
\label{tab:road-graph-eval}
\setlength\tabcolsep{9.0pt}
\begin{threeparttable}
    \begin{tabular}{l|ccc}
        \toprule 
        \textbf{Agent count} & \textbf{Precision} [\unit{\percent}] & \textbf{Recall} [\unit{\percent}] & \textbf{F\textsubscript{1}-score} [\unit{\percent}] \\
        \midrule
        \multicolumn{4}{l}{\textit{OSM intersections (car did not turn):}} \\
        [.5ex]
        1 agent  &         85.52 &	50.72 &	63.68 \\
        2 agents &         83.94 &	52.90 &	64.90 \\ 
        3 agents &         83.33 &	54.35 &	65.79 \\
        \midrule
        \multicolumn{4}{l}{\textit{OSM intersections (car turned):}} \\
        [.5ex]
        1 agent  & \underline{76.75} & 91.30 & 83.40 \\
        2 agents & \textbf{77.03} &	\underline{97.10} & \underline{85.91} \\
        3 agents & 76.67 & \textbf{100.00}\phantom{0} & \textbf{86.79} \\
        \bottomrule
    \end{tabular}
    Evaluation of the estimated intersections contained in the road graph layer. The ground truth intersections are obtained from OpenStreetMap (OSM). While the upper block contains all OSM intersections, the lower block refers to those that could be detected by our approach (see \cref{fig:osm-gt}). Bold and underlined values indicate the best and second-best scores, respectively.
\end{threeparttable}
\vspace*{-.4cm}
\end{table}

In the final experiment, we assess the effectiveness of our heuristic-based approach for detecting intersections within the road graph.
As a reference, we extract the location of all intersections along the driven route from OpenStreetMap~(OSM). Specifically, we define an intersection as the junction of at least two roads with different street names. We merge spatially proximate intersections and retain the averaged location in the reference set if the distance to the traveled route is less than \qty{12}{\meter}. 
As discussed in \cref{sec:approach-scene-graph}, our approach can only detect intersections when an agent executes a turn. Therefore, we manually curate a subset of reference intersections that fall within the scope of our method. We visualize both the complete reference set (in red) and the filtered subset (in green) in \cref{fig:osm-gt}.

For evaluation, we pair each estimated intersection with the nearest intersection in the OSM reference set and consider a match as correct if the distance is below \qty{50}{\meter}. For reference, the average length of a single agent's route is approximately \qty{10}{\km}, within a total area of \qty{1.3}{\km\squared}. We report the precision, recall, and F\textsubscript{1}-score for both reference sets in \cref{tab:road-graph-eval}.
First, intersections within the scope of our approach are detected with high certainty, as indicated by the recall values in the bottom block. Therefore, we hypothesize that a broader coverage of the environment would likely improve the recall for the complete reference set.
Second, the precision values in the bottom block are lower than the corresponding recall, which we attribute to the approach’s inability to distinguish intersections from sharp turns.
Finally, the performance remains relatively stable across different agent counts, demonstrating the effective integration of multi-agent information into the scene graph.
\section{Conclusion}
In this work, we introduced CURB-OSG, a novel approach for building hierarchical, dynamic 3D scene graphs of large-scale urban driving environments from multiple observing agents. Our approach addresses the key limitations of previous methods by directly processing real-world sensor data, leveraging open-vocabulary semantics, and operating independently of initial agent pose estimates. Despite the increased complexity, our approach enhances the robustness of object proposal fusion. To foster further advancements in this domain, we make our code publicly available. Future work will focus on improving dynamic object tracking and incorporating semantic loop closure by leveraging identified scene graph entities.



\footnotesize
\bibliographystyle{IEEEtran}

\begin{thebibliography}{10}
\providecommand{\url}[1]{#1}
\csname url@rmstyle\endcsname
\providecommand{\newblock}{\relax}
\providecommand{\bibinfo}[2]{#2}
\providecommand\BIBentrySTDinterwordspacing{\spaceskip=0pt\relax}
\providecommand\BIBentryALTinterwordstretchfactor{4}
\providecommand\BIBentryALTinterwordspacing{\spaceskip=\fontdimen2\font plus
\BIBentryALTinterwordstretchfactor\fontdimen3\font minus \fontdimen4\font\relax}
\providecommand\BIBforeignlanguage[2]{{%
\expandafter\ifx\csname l@#1\endcsname\relax
\typeout{** WARNING: IEEEtran.bst: No hyphenation pattern has been}%
\typeout{** loaded for the language `#1'. Using the pattern for}%
\typeout{** the default language instead.}%
\else
\language=\csname l@#1\endcsname
\fi
#2}}

\bibitem{greve2024curbsg}
E.~Greve, M.~Büchner, N.~Vödisch, W.~Burgard, and A.~Valada, ``Collaborative dynamic {3D} scene graphs for automated driving,'' in \emph{{IEEE} International Conference on Robotics and Automation}, 2024, pp. 11\,118--11\,124.

\bibitem{hovsg}
A.~Werby, C.~Huang, M.~Büchner, A.~Valada, and W.~Burgard, ``Hierarchical open-vocabulary {3D} scene graphs for language-grounded robot navigation,'' in \emph{Robotics: Science and Systems}, 2024.

\bibitem{buchner2023learning}
M.~B{\"u}chner, J.~Z{\"u}rn, I.-G. Todoran, A.~Valada, and W.~Burgard, ``Learning and aggregating lane graphs for urban automated driving,'' in \emph{{IEEE/CVF} Conf.~on Computer Vision and Pattern Recognition}, 2023, pp. 13\,415--13\,424.

\bibitem{honerkamp2024language}
D.~Honerkamp, M.~Büchner, F.~Despinoy, T.~Welschehold, and A.~Valada, ``Language-grounded dynamic scene graphs for interactive object search with mobile manipulation,'' \emph{{IEEE} Robotics and Automation Letters}, vol.~9, no.~10, pp. 8298--8305, 2024.

\bibitem{deng_opengraph_2024}
Y.~Deng, J.~Wang, J.~Zhao, X.~Tian, G.~Chen, Y.~Yang, and Y.~Yue, ``{OpenGraph}: Open-vocabulary hierarchical {3D} graph representation in large-scale outdoor environments,'' \emph{{IEEE} Robotics and Automation Letters}, vol.~9, no.~10, pp. 8402--8409, 2024.

\bibitem{hughes2022hydra}
N.~Hughes, Y.~Chang, and L.~Carlone, ``Hydra: A real-time spatial perception system for {3D} scene graph construction and optimization,'' in \emph{Robotics: Science and Systems}, 2022.

\bibitem{gu_conceptgraphs_2024}
Q.~Gu, A.~Kuwajerwala, S.~Morin, K.~M. Jatavallabhula, B.~Sen, A.~Agarwal, \emph{et~al.}, ``{ConceptGraphs}: Open-vocabulary {3D} scene graphs for perception and planning,'' in \emph{{IEEE} International Conference on Robotics and Automation}, 2024, pp. 5021--5028.

\bibitem{chang_hydra-multi_2023}
Y.~Chang, N.~Hughes, A.~Ray, and L.~Carlone, ``{Hydra-Multi}: Collaborative online construction of {3D} scene graphs with multi-robot teams,'' in \emph{{IEEE/RSJ} International Conference on Intelligent Robots and Systems}, 2023, pp. 10\,995--11\,002.

\bibitem{DosovitskiyCarla17}
A.~Dosovitskiy, G.~Ros, F.~Codevilla, A.~Lopez, and V.~Koltun, ``{CARLA}: {An} open urban driving simulator,'' in \emph{Conference on Robot Learning}, 2017.

\bibitem{barnes2020robotcarradar}
D.~Barnes, M.~Gadd, P.~Murcutt, P.~Newman, and I.~Posner, ``{The Oxford Radar RobotCar Dataset}: A radar extension to the {Oxford RobotCar Dataset},'' in \emph{{IEEE} International Conference on Robotics and Automation}, 2020, pp. 6433--6438.

\bibitem{zhang2014loam}
J.~Zhang and S.~Singh, ``{LOAM}: Lidar odometry and mapping in real-time,'' in \emph{Robotics: Science and Systems}, 2014.

\bibitem{vizzo2023kissicp}
I.~Vizzo, T.~Guadagnino, B.~Mersch, L.~Wiesmann, J.~Behley, and C.~Stachniss, ``{KISS-ICP}: In defense of point-to-point {ICP} – simple, accurate, and robust registration if done the right way,'' \emph{{IEEE} Robotics and Automation Letters}, vol.~8, no.~2, pp. 1029--1036, 2023.

\bibitem{koide2019hdlgraphslam}
K.~Koide, J.~Miura, and E.~Menegatti, ``A portable three-dimensional {LiDAR}-based system for long-term and wide-area people behavior measurement,'' \emph{International Journal of Advanced Robotic Systems}, vol.~16, no.~2, 2019.

\bibitem{kim2018scancontext}
G.~Kim and A.~Kim, ``{Scan Context}: Egocentric spatial descriptor for place recognition within {3D} point cloud map,'' in \emph{{IEEE/RSJ} International Conference on Intelligent Robots and Systems}, 2018, pp. 4802--4809.

\bibitem{li2021semanticscancontext}
L.~Li, X.~Kong, X.~Zhao, T.~Huang, W.~Li, F.~Wen, H.~Zhang, and Y.~Liu, ``{SSC}: {Semantic Scan Context} for large-scale place recognition,'' in \emph{{IEEE/RSJ} International Conference on Intelligent Robots and Systems}, 2021, pp. 2092--2099.

\bibitem{voedisch2025vfmreg}
N.~Vödisch, G.~Cioffi, M.~Cannici, W.~Burgard, and D.~Scaramuzza, ``{LiDAR} registration with visual foundation models,'' \emph{arXiv preprint arXiv:2502.19374}, 2025.

\bibitem{arce2023padloc}
J.~Arce, N.~Vödisch, D.~Cattaneo, W.~Burgard, and A.~Valada, ``{PADLoC}: {LiDAR}-based deep loop closure detection and registration using panoptic attention,'' \emph{{IEEE} Robotics and Automation Letters}, vol.~8, no.~3, pp. 1319--1326, 2023.

\bibitem{cattaneo2022lcdnet}
D.~Cattaneo, M.~Vaghi, and A.~Valada, ``{LCDNet}: Deep loop closure detection and point cloud registration for {LiDAR SLAM},'' \emph{{IEEE} Robotics and Automation Letters}, vol.~38, no.~4, pp. 2074--2093, 2022.

\bibitem{zou2019collaborative}
D.~Zou, P.~Tan, and W.~Yu, ``Collaborative visual {SLAM} for multiple agents: A brief survey,'' \emph{Virtual Reality \& Intelligent Hardware}, vol.~1, no.~5, pp. 461--482, 2019.

\bibitem{riazuelo2014c2tam}
L.~Riazuelo, J.~Civera, and J.~Montiel, ``{C\textsuperscript{2}TAM}: A cloud framework for cooperative tracking and mapping,'' \emph{Robotics and Autonomous Systems}, vol.~62, no.~4, pp. 401--413, 2014.

\bibitem{Karrer2018cvislam}
M.~Karrer, P.~Schmuck, and M.~Chli, ``{CVI-SLAM} — collaborative visual-inertial {SLAM},'' \emph{{IEEE} Robotics and Automation Letters}, vol.~3, no.~4, pp. 2762--2769, 2018.

\bibitem{chang2022lamp2}
Y.~Chang, K.~Ebadi, C.~E. Denniston, M.~F. Ginting, A.~Rosinol, A.~Reinke, \emph{et~al.}, ``{LAMP 2.0}: A robust multi-robot {SLAM} system for operation in challenging large-scale underground environments,'' \emph{{IEEE} Robotics and Automation Letters}, vol.~7, no.~4, pp. 9175--9182, 2022.

\bibitem{lajoie2024swarmslam}
P.-Y. Lajoie and G.~Beltrame, ``{Swarm-SLAM}: Sparse decentralized collaborative simultaneous localization and mapping framework for multi-robot systems,'' \emph{{IEEE} Robotics and Automation Letters}, vol.~9, no.~1, pp. 475--482, 2024.

\bibitem{huang2022discoslam}
Y.~Huang, T.~Shan, F.~Chen, and B.~Englot, ``{DiSCo-SLAM}: Distributed scan context-enabled multi-robot {LiDAR SLAM} with two-stage global-local graph optimization,'' \emph{{IEEE} Robotics and Automation Letters}, vol.~7, no.~2, pp. 1150--1157, 2022.

\bibitem{armeni20193d}
I.~Armeni, Z.-Y. He, A.~Zamir, J.~Gwak, J.~Malik, M.~Fischer, and S.~Savarese, ``{3D} scene graph: A structure for unified semantics, {3D} space, and camera,'' in \emph{International Conference on Computer Vision}, 2019, pp. 5663--5672.

\bibitem{sgraphs_2022}
H.~Bavle, J.~L. Sanchez-Lopez, M.~Shaheer, J.~Civera, and H.~Voos, ``Situational graphs for robot navigation in structured indoor environments,'' \emph{{IEEE} Robotics and Automation Letters}, vol.~7, no.~4, pp. 9107--9114, 2022.

\bibitem{wu_scenegraphfusion}
S.-C. Wu, J.~Wald, K.~Tateno, N.~Navab, and F.~Tombari, ``{SceneGraphFusion}: Incremental {3D} scene graph prediction from {RGB-D} sequences,'' in \emph{{IEEE/CVF} Conf.~on Computer Vision and Pattern Recognition}, June 2021, pp. 7515--7525.

\bibitem{voedisch2025pastel}
N.~Vödisch, K.~Petek, M.~Käppeler, A.~Valada, and W.~Burgard, ``A good foundation is worth many labels: Label-efficient panoptic segmentation,'' \emph{{IEEE} Robotics and Automation Letters}, vol.~10, no.~1, pp. 216--223, 2025.

\bibitem{redmonYouOnlyLook2016}
J.~Redmon, S.~Divvala, R.~Girshick, and A.~Farhadi, ``You only look once: Unified, real-time object detection,'' in \emph{{IEEE/CVF} Conf.~on Computer Vision and Pattern Recognition}, 2016, pp. 779--788.

\bibitem{liGroundedLanguageImagePretraining2022}
L.~H. Li, P.~Zhang, H.~Zhang, J.~Yang, C.~Li, Y.~Zhong, \emph{et~al.}, ``Grounded language-image pre-training,'' in \emph{{IEEE/CVF} Conf.~on Computer Vision and Pattern Recognition}, 2022, pp. 10\,955--10\,965.

\bibitem{liBLIPBootstrappingLanguageImage2022a}
J.~Li, D.~Li, C.~Xiong, and S.~Hoi, ``{BLIP}: Bootstrapping language-image pre-training for unified vision-language understanding and generation,'' in \emph{International Conference on Machine Learning}, 2022, pp. 12\,888--12\,900.

\bibitem{kirillovSegmentAnything2023}
A.~Kirillov, E.~Mintun, N.~Ravi, H.~Mao, C.~Rolland, L.~Gustafson, \emph{et~al.}, ``Segment anything,'' in \emph{International Conference on Computer Vision}, 2023, pp. 4015--4026.

\bibitem{liuGroundingDINOMarrying2024}
S.~Liu, Z.~Zeng, T.~Ren, F.~Li, H.~Zhang, J.~Yang, \emph{et~al.}, ``{Grounding DINO}: Marrying {DINO} with grounded pre-training for open-set object detection,'' in \emph{European Conference on Computer Vision}, 2024, pp. 38--55.

\bibitem{renGroundedSAMAssembling2024}
T.~Ren, S.~Liu, A.~Zeng, J.~Lin, K.~Li, H.~Cao, J.~Chen, X.~Huang, Y.~Chen, F.~Yan, \emph{et~al.}, ``{Grounded SAM}: Assembling open-world models for diverse visual tasks,'' \emph{arXiv preprint arXiv:2401.14159}, 2024.

\bibitem{zhangRecognizeAnythingStrong2024}
Y.~Zhang, X.~Huang, J.~Ma, Z.~Li, Z.~Luo, Y.~Xie, \emph{et~al.}, ``Recognize anything: A strong image tagging model,'' in \emph{{IEEE/CVF} Conference on Computer Vision and Pattern Recognition Workshops}, 2024, pp. 1724--1732.

\bibitem{strader_indoor_2024}
J.~Strader, N.~Hughes, W.~Chen, A.~Speranzon, and L.~Carlone, ``Indoor and outdoor {3D} scene graph generation via language-enabled spatial ontologies,'' \emph{{IEEE} Robotics and Automation Letters}, vol.~9, no.~6, pp. 4886--4893, 2024.

\bibitem{liMatchingAnythingSegmenting2024}
S.~Li, L.~Ke, M.~Danelljan, L.~Piccinelli, M.~Segu, L.~V. Gool, and F.~Yu, ``Matching anything by segmenting anything,'' in \emph{{IEEE/CVF} Conf.~on Computer Vision and Pattern Recognition}, 2024, pp. 18\,963--18\,973.

\bibitem{koideVoxelizedGICPFast2021}
K.~Koide, M.~Yokozuka, S.~Oishi, and A.~Banno, ``Voxelized {GICP} for fast and accurate {3D} point cloud registration,'' in \emph{{IEEE} International Conference on Robotics and Automation}, 2021, pp. 11\,054--11\,059.

\bibitem{kuemmerle2011g2o}
R.~Kümmerle, G.~Grisetti, H.~Strasdat, K.~Konolige, and W.~Burgard, ``{g\textsuperscript{2}o}: A general framework for graph optimization,'' in \emph{{IEEE} Robotics and Automation Letters}, 2011, pp. 3607--3613.

\bibitem{cordts2016cityscapes}
M.~Cordts, M.~Omran, S.~Ramos, T.~Rehfeld, M.~Enzweiler, R.~Benenson, U.~Franke, S.~Roth, and B.~Schiele, ``The {Cityscapes} dataset for semantic urban scene understanding,'' in \emph{{IEEE/CVF} Conf.~on Computer Vision and Pattern Recognition}, 2016, pp. 3213--3223.

\bibitem{pan2024tokenizeAnything}
T.~Pan, L.~Tang, X.~Wang, and S.~Shan, ``Tokenize anything via prompting,'' in \emph{European Conference on Computer Vision}, 2024, pp. 330--348.

\bibitem{ester1996dbscan}
M.~Ester, H.-P. Kriegel, J.~Sander, X.~Xu, \emph{et~al.}, ``A density-based algorithm for discovering clusters in large spatial databases with noise,'' in \emph{{ACM SIGKDD} Conference on Knowledge Discovery and Data Mining}, 1996, pp. 226--231.

\bibitem{filipenko2018comparison}
M.~Filipenko and I.~Afanasyev, ``Comparison of various {SLAM} systems for mobile robot in an indoor environment,'' in \emph{{IEEE} International Conference on Intelligent Systems}, 2018, pp. 400--407.

\bibitem{geiger2013vision}
A.~Geiger, P.~Lenz, C.~Stiller, and R.~Urtasun, ``Vision meets robotics: The {KITTI} dataset,'' \emph{International Journal of Robotics Research}, vol.~32, no.~11, pp. 1231--1237, 2013.

\bibitem{chen2019iros}
X.~Chen, A.~Milioto, E.~Palazzolo, P.~Giguère, J.~Behley, and C.~Stachniss, ``{SuMa++}: Efficient {LiDAR}-based semantic {SLAM},'' in \emph{{IEEE/RSJ} International Conference on Intelligent Robots and Systems}, 2019, pp. 4530--4537.

\bibitem{zhou2022maskclip}
C.~Zhou, C.~C. Loy, and B.~Dai, ``Extract free dense labels from {CLIP},'' in \emph{European Conference on Computer Vision}, 2022, pp. 696--712.

\bibitem{fufeatup}
S.~Fu, M.~Hamilton, L.~E. Brandt, A.~Feldmann, Z.~Zhang, and W.~T. Freeman, ``{FeatUp}: A model-agnostic framework for features at any resolution,'' in \emph{International Conf.~on Learning Representations}, 2024.

\bibitem{he2017connectedComponents}
L.~He, X.~Ren, Q.~Gao, X.~Zhao, B.~Yao, and Y.~Chao, ``The connected-component labeling problem: A review of state-of-the-art algorithms,'' \emph{Pattern Recognition}, pp. 25--43, 2017.

\end{thebibliography}



\end{document}